\documentclass[journal]{IEEEtran}
\usepackage[utf8]{inputenc}
\usepackage{graphicx}
\usepackage[table,x11names]{xcolor}
\usepackage{pifont}
\usepackage{utfsym}
\usepackage{amsmath,amsfonts}
\usepackage{algorithm}
\usepackage{algorithmic}
\usepackage{stfloats}
\usepackage[caption=false,font=normalsize,labelfont=sf,textfont=sf]{subfig}
\usepackage{booktabs}
\usepackage{array}
\usepackage{multirow}
\usepackage{arydshln}

\usepackage[switch]{lineno} 

\usepackage{microtype}
\usepackage{textcomp}
\usepackage{enumitem}
\usepackage{url}
\usepackage{verbatim}
\usepackage[pagebackref=true,breaklinks=true,colorlinks,bookmarks=false,urlcolor=red,citecolor=cyan]{hyperref}
\begin{document}
\title{PolSAM: \textbf{Pol}arimetric Scattering Mechanism Informed \textbf{S}egment \textbf{A}nything \textbf{M}odel}

\author{Yuqing Wang, Zhongling Huang,~\IEEEmembership{Member,~IEEE,} Shuxin Yang, Hao Tang, Xiaolan Qiu,~\IEEEmembership{Senior Member,~IEEE,} Junwei Han,~\IEEEmembership{Fellow,~IEEE,} and Dingwen Zhang,~\IEEEmembership{Member,~IEEE}
\thanks{This work was supported by the National Natural Science Foundation of China (Grant No. 62571443 and Grant No. 62293543); the National Key Laboratory of Microwave Imaging Technology; the Natural Science Basic Research Program of Shaanxi (Program No. 2025JC-QYXQ-032); and the Guangdong Basic and Applied Basic Research Foundation (Grant No. 2025A1515011368). (Corresponding authors: Zhongling Huang; Dingwen Zhang.)}

\thanks{Yuqing Wang, Zhongling Huang, Shuxin Yang, Junwei Han and Dingwen Zhang are with Brain and Artificial Intelligence Laboratory, School of Automation, Northwestern Polytechnical University, Xi’an, 710072, China, and also with the Shenzhen Research Institute of Northwestern Polytechnical University, Shenzhen 518057, China (e-mail: wanda33683@gmail.com, huangzhongling@nwpu.edu.cn, yangshuxin@mail.nwpu.edu.cn, junweihan2010@gmail.com, zhangdingwen2006yyy@gmail.com).}
\thanks{Hao Tang is with the National Key Laboratory for Multimedia Information Processing, School of Computer Science, Peking University, Beijing 100871, China. (e-mail: haotang@pku.edu.cn).}
\thanks{Xiaolan Qiu is with the National Key Laboratory of Microwave Imaging Technology, Chinese Academy of Sciences, Beijing 100090, China, and also with the Aerospace Information Research Institute, Chinese Academy of Sciences, Beijing 100094, China (e-mail: xlqiu@mail.ie.ac.cn).}


}

\markboth{Journal of \LaTeX\ Class Files,~Vol.~14, No.~8, August~2021}%
{Shell \MakeLowercase{\textit{et al.}}: A Sample Article Using IEEEtran.cls for IEEE Journals}

\maketitle

\begin{abstract}
PolSAR data presents unique challenges due to its rich and complex characteristics. Existing data representations, such as complex-valued data, polarimetric features, and amplitude images, are widely used. However, these formats often face issues related to usability, interpretability, and data integrity. While most feature extraction networks for PolSAR attempt to address these issues, they are typically small in size, which limits their ability to effectively extract features. To overcome these limitations, we introduce the large, powerful Segment Anything Model (SAM), which excels in feature extraction and prompt-based segmentation. However, SAM's application to PolSAR is hindered by modality differences and limited integration of domain-specific knowledge. 
To address this, we propose the Polarimetric Scattering Mechanism-Informed SAM (PolSAM). Central to our approach is the Microwave Vision Data (MVD) representation, which distills complex scattering mechanisms into a lightweight, physically interpretable format. We integrate MVD into the frozen SAM encoder via a Feature-Level Fusion Prompt (FFP) module and specialized adapters, effectively bridging the modality gap by merging pseudo-colored SAR features with physical cues. Furthermore, a Semantic-Level Fusion Prompt (SFP) module is designed to progressively refine segmentation through sparse and dense prompt embeddings derived from MVD's semantic correlations. Experiments on the newly constructed PhySAR-Seg datasets demonstrate that PolSAM significantly outperforms existing SAM-based and multimodal fusion methods, achieving superior segmentation accuracy with enhanced data storage and inference efficiency.
The source code and data will be publicly available at \url{https://github.com/XAI4SAR/PolSAM}.
\end{abstract}

\begin{figure}[!t] 
    \centering
    \includegraphics[width=0.99\columnwidth]{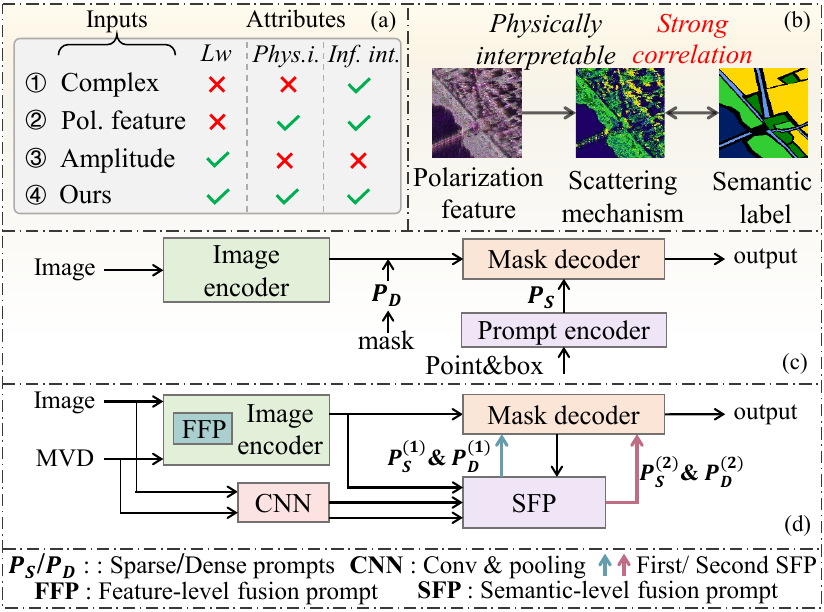}
    \caption{(a) Different input representations of PolSAR data and their corresponding attributes, including lightweight (Lw), physical interpretability (Phys. i.), and information integrity (Inf. int.), are explored. (b) Scattering mechanisms depict polarization characteristics with physical interpretability and exhibit a strong correlation with semantic information. (c) Original SAM architecture; (d) Our PolSAM model.}
    \label{fig::intro}
\end{figure}

\section{Introduction}

\IEEEPARstart{P}{olarimetric} Synthetic Aperture Radar (PolSAR) provides rich polarization features, enabling detailed terrain and ground target representation for applications like land-cover segmentation with enhanced accuracy and feature discrimination \cite{bi2020polarimetric, zhang2025ccsfuse, zhang2025unsupervised}. Recently, deep learning-based methods for PolSAR image segmentation have gained traction, typically reformatting PolSAR data into specific input types and designing tailored neural networks to extract both image and polarimetric features \cite{zhang2020unsupervised,qin2020polsar,li2025multiscale, li2026langsurf}. Common representations include complex-valued inputs \cite{li2020complex}, model-based polarimetric features \cite{antropov2013land,shen2020residual,ghanbari2023local}, and amplitude images \cite{pu2025classwise}, processed using architectures like complex-valued neural networks (CVNNs) \cite{yu2022lightweight, kuang2024polarimetry}, convolutional neural networks (CNNs) \cite{ghanbari2023local, zhang2020unsupervised, chen2025enhanced}, Transformers \cite{hua2024feature, fang2025hybrid}, and graph neural networks (GNNs) \cite{dong2021exploring, zhang2025unsupervised, xu2025lightweight}. However, these input formats face challenges in usability, interpretability, and data integrity, while these models are relatively small, limiting their representational capacity. 

As illustrated in Fig. \ref{fig::intro}(a), complex-valued inputs retain the complete information of PolSAR data but require significantly high storage capacity. Furthermore, their processing necessitates the use of specialized CVNNs, which introduces additional computational complexity and weakens interpretability. This constraint also limits the applicability of existing large-scale vision models, as they are typically designed to handle real-valued data, making the integration of complex-valued inputs more challenging. Model-based polarimetric features enhance the interpretability of complex-valued PolSAR data, enabling effective integration with popular deep learning architectures like CNNs, Transformers, and GNNs. However, these features remain inherently high-dimensional. The most widely used format is the 8-bit quantized amplitude PolSAR image, aligned with photographic conventions for compatibility with existing deep learning models \cite{zhang2020exploring, liu2022disentangled, zhang2023weakly}. Nevertheless, this approach sacrifices the rich polarimetric features, leading to significant information loss. Investigating a user-friendly input format for PolSAR data that retains integrated polarimetric information is vital in deep learning applications.

In PolSAR image segmentation, the challenges discussed above are further intensified by the limited availability of data. This limitation is particularly pronounced in networks with fewer parameters, as they often fail to extract sufficient information from the input data. A common solution is to transfer knowledge from pre-trained models \cite{wu2019polsar}, but the high dimensionality of polarimetric features often makes them incompatible with standard image representations \cite{huang2018supervised}. Although CVNNs have twice the number of parameters compared to traditional real-valued networks, the lack of effective pre-trained models limits their utility. In data-scarce scenarios, CVNNs often struggle to converge, further undermining their practical applicability \cite{zeng2022ts, zeng2023semipscn}. While various advanced architectures have been proposed to mitigate these challenges, the inherent limitations of small network sizes hinder their ability to effectively extract relevant features. Additionally, the diversity of PolSAR datasets, variations in scattering mechanisms, and differences in imaging conditions complicate the development of models capable of delivering robust performance across diverse scenarios \cite{ghanbari2023local, tong2025terrain}. While recent methods address scattering ambiguities and environmental variations through adaptive superpixel aggregation \cite{li2025multiscale}, composite physical descriptors \cite{xiang2025oil}, temporal constraints \cite{li2025edge}, amplitude-invariant representations \cite{li2025caifnet}, or azimuth-sensitive PolSAR image generation \cite{xie2026azi}, PolSAM tackles these challenges by synergizing the macroscopic stability of MVD with the robust contextual reasoning of SAM.

To address these challenges, this work focuses on designing an efficient input format for PolSAR data that integrates Pauli pseudo-colored images with inherent scattering properties of PolSAR, preserving the core polarimetric features crucial for analysis. Previous studies \cite{hua2024feature, huang2022physically} have demonstrated the critical role of scattering mechanisms in PolSAR analysis. As shown in Fig. \ref{fig::intro}(b), we optimize the scattering properties into a compact image format, reducing storage requirements while providing intuitive physical interpretability. Moreover, specific scattering patterns basically correspond to particular land cover categories, highlighting a strong correlation with semantic label information. With this foundation, we aim to create a generalized segmentation framework for PolSAR image that leverages the robust capabilities of existing large vision models. This approach seeks to bridge the gap between the unique characteristics of PolSAR data and the good generalization ability of advanced large vision models, enabling broader applications and improved usability for non-expert users.

Building on these insights, we introduce the Segment Anything Model (SAM) \cite{kirillov2023segment}, as shown in Fig. \ref{fig::intro}(c). First, SAM leverages its large-scale architecture and robust data processing capabilities to achieve efficient and precise segmentation across diverse image content, making it a promising pre-trained foundation model for PolSAR applications. Second, SAM's prompt-based segmentation mechanism effectively incorporates high-level external prompts, guiding the model to focus on relevant features. This capability is particularly well-suited for compact, high-level scattering mechanism data, providing semantic guidance to enhance segmentation accuracy.

Overall, we propose the Polarimetric Scattering Mechanism-Informed SAM (PolSAM), a framework built upon a novel data representation termed Microwave Vision Data (MVD). MVD serves as a lightweight, physically interpretable, and information-dense encoding of scattering mechanism classifications. To integrate physical scattering properties into SAM, we use two key modules, as shown in Fig. \ref{fig::intro}(d). The Feature-Level Fusion Prompt (FFP) module first merges patch embeddings from the Pauli decomposition-based pseudo-colored image and MVD, enhancing their interaction before feeding into the SAM encoder. Dedicated adapters in each encoder layer address modality incompatibility in the frozen SAM encoder, fostering feature complementarity. The Semantic-Level Fusion Prompt (SFP) module, second, refines semantic guidance for segmentation through a two-stage process. First, it integrates input features with encoder outputs; then, it aligns the fused representations with high-level semantic prompts, boosting segmentation performance. 

Our contributions are as follows:
\begin{enumerate}
\item{We propose a novel pipeline for PolSAR data that generates user-friendly, physically interpretable, and information-dense MVD products. These lightweight products leverage the strong correlation between scattering mechanisms and semantics to aid in terrain segmentation.}
\item{We propose PolSAM, a SAM-based PolSAR segmentation method that combines PolSAR domain-specific knowledge with SAM’s strengths to improve efficiency and accuracy. PolSAM includes FFP, which fuses Pauli pseudo-colored images and MVD embeddings for better feature complementarity, and SFP, which uses a two-level design to refine semantic prompts through enhanced input and high-level semantic feature interaction.}
\item{Experiments on the PhySAR-Seg datasets show that PolSAM outperforms SAM-based and multimodal fusion models. The sparse and dense prompts effectively capture semantic relevance, guiding segmentation tasks. Additionally, MVD integration reduces storage requirements, offering a more efficient alternative to traditional polarimetric representations.}
\end{enumerate}

\section{Related Works}
\label{sec:related work}
\subsection{Deep learning based PolSAR image analysis}
Deep learning techniques have become essential in PolSAR image analysis, leveraging their advanced information processing capabilities to improve segmentation performance \cite{qin2020polsar,ghanbari2023local}. Currently, widely used methods include CNNs or Transformer-based architectures, GNNs, and CVNNs. The inputs processed by these methods can be broadly classified into three categories: complex-valued data, model-based polarimetric features, and simple amplitude data. 

As shown in Fig. \ref{fig::intro}(a), "\textcolor{green}{\ding{51}}" denotes an affirmative indicator, "\textcolor{red}{\ding{53}}" denotes a negative indicator. Complex-valued data is mainly processed using CVNNs \cite{li2020complex, yu2022lightweight, kuang2024polarimetry}. Clearly, this approach satisfies information integrity (Inf. int.), but it requires high storage and computational resources, making it not lightweight (Lw), while also lacking physical interpretability (Phys. i.). Polarimetric features (Pol. feature), which mainly include Pauli decomposition-based features \cite{antropov2013land}, polarization coherence matrix \( T \) \cite{shen2020residual,gao2023dualistic,zhang2023learning}, polarization covariance matrix \( C \) \cite{ghanbari2023local,xiang2024polsar}, and scattering features extracted from traditional polarization decomposition methods such as Cloude-Pottier, H/A/Alpha decomposition, Freeman-Durden three-component decomposition, and Yamaguchi four-component decomposition, provide physically interpretable and relatively rich information. These features enhance the ability of deep learning models to interpret PolSAR data \cite{xiao2020terrain,shi2023cnn} . However, they are remain high-dimensional, resulting in high storage and computational costs, making them not lightweight. In comparison, amplitude data, typically represented as 8-bit quantized grayscale images or pseudo-colored images visualized using different polarization modes (HH, HV, VH), is lightweight but suffers from significant information loss and lacks physical interpretability \cite{pu2025classwise}. In summary, these methods face challenges in simultaneously achieving lightweight representation, physical interpretability, and information richness.

Polarimetric features and amplitude data formats can both be processed using networks such as CNNs, Transformers, and GNNs. CNNs and Transformers aim to capture fundamental structural and texture details by enhancing polarization feature representation through advanced techniques \cite{zhang2020unsupervised, hua2024feature, chen2025enhanced, zhang2023learning, shi2024content}. GNNs optimize spatial relationships in PolSAR images by modeling complex spatial dependencies and integrating multi-scale information \cite{geng2022polarimetric, shi2023cnn, wang2024multiscale, xu2025statistic}. Despite the aforementioned advancements, these networks often require task-specific designs and their limited scale restricts scalability and computational efficiency.

It is worth noting that existing deep learning models do not directly process the classification results of scattering mechanisms but instead focus on scattering features extracted by traditional decomposition methods. The MVD product we propose represents the classification results of scattering mechanisms, providing higher-level physical information that not only offers better physical interpretability but also has a strong correlation with semantic information. As shown in Fig. \ref{fig::intro}(a), ours, which combines MVD with Pauli pseudo-colored images, achieves a lightweight, user-friendly representation while preserving physical interpretability and information integrity.

\subsection{SAM Implementation in Segmentation}
The SAM model has demonstrated considerable success in segmentation tasks, primarily due to its large model size, which enables powerful feature extraction and broad applicability. Additionally, its flexible prompt-based segmentation mechanism allows it to effectively handle external high-level prompt information \cite{zhang2024personalize, zhang2023faster, wu2025medical, chai2024ladder}. This versatility has driven adoption across diverse domains, including natural images \cite{zhang2023faster, zhang2024personalize, ke2024segment}, medical imaging \cite{yue2024surgicalsam, ali2025review, xu2024esp, wu2025trans, xie2025self}, and remote sensing \cite{zhang2025rsam, chen2024rsprompter, yan2023ringmo, zheng2024tuning, zhang2025alps}. For instance, recent studies have systematically benchmarked and unleashed the capabilities of generic segmentation models for complex remote sensing imagery \cite{zhang2024unleashing}. However, direct application often struggles with inherent modality differences, necessitating tailored approaches. To bridge these gaps, researchers have adapted SAM using specialized adapter techniques \cite{wu2025medical, zhang2025rsam, pu2025classwise} or designed modality-specific refinements. For instance, Classwise-SAM-Adapter \cite{pu2025classwise} transfers SAM to SAR land-cover segmentation, yet without involving polarimetric information. Another notable example is IRSAM \cite{zhang2024irsam}, which successfully adapts the framework to the challenging infrared domain by addressing low-contrast features. Beyond structural adaptations, techniques such as integrating text prompts within segmentation modules have also enhanced performance \cite{yue2023surgicalpart}. Such multimodal strategies are more prevalent in natural image analysis, where innovations explore the use of a multi-modal encoder that combines visual and linguistic information to generate prompt embeddings \cite{zhang2026evf,liu2025refersam, tang2025ssp}. Furthermore, PA-SAM \cite{xie2024pa} introduces a prompt-driven adapter to enhance mask quality by refining sparse and dense prompt features. Altogether, these adaptations highlight SAM’s capacity for domain-specific adjustments, extending its applications across diverse image types.

PolSAR data's scattering mechanism classification results are high-level compared to polarimetric and scattering features, and SAM's strengths are well-suited to address the challenges we previously outlined. However, despite many of the above methods effectively adjusting SAM, it still faces challenges when applied to PolSAR images, mainly due to modality differences and the inability to integrate domain-specific knowledge. For this, our model introduces two specialized prompt learning modules: one enhances feature complementarity through early-stage fusion of SAR and MVD embeddings, while the other generates precise sparse and dense prompts through a two-level fusion of features and semantic context, effectively guiding the decoder in the segmentation process.
\begin{figure}[!htbp]
    \centering 
    \includegraphics[width=0.99\columnwidth]{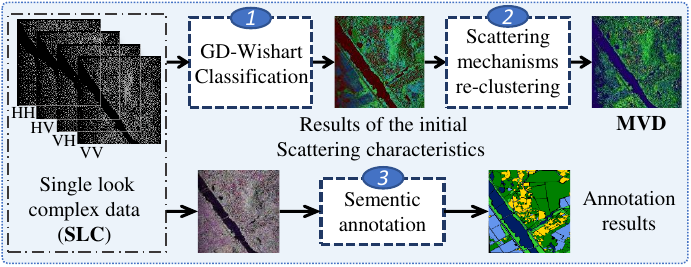}
    \caption{Data processing pipeline: unsupervised GD-Wishart classification, MVD generation through re-clustering of scattering mechanisms, and semantic annotation, illustrated with the PhySAR-Seg-1 dataset.} 
    \label{fig::data_process}
\end{figure}
\section{PhySAR-Seg Dataset}
To tackle the data challenges mentioned above, we propose a novel pipeline for constructing the PhySAR-Seg dataset, which comprises two subsets, PhySAR-Seg-1 and PhySAR-Seg-2. The key innovation is Microwave Vision Data (MVD), a lightweight and user-friendly representation that ensures high storage efficiency and provides intuitive, physically interpretable visualizations of scattering mechanisms. By integrating MVD with Pauli pseudo-colored images, we also ensure information integrity. Details of the pipeline and dataset are provided below.

\begin{figure*}[!htbp]
    \centering 
    \includegraphics[width=1\textwidth]{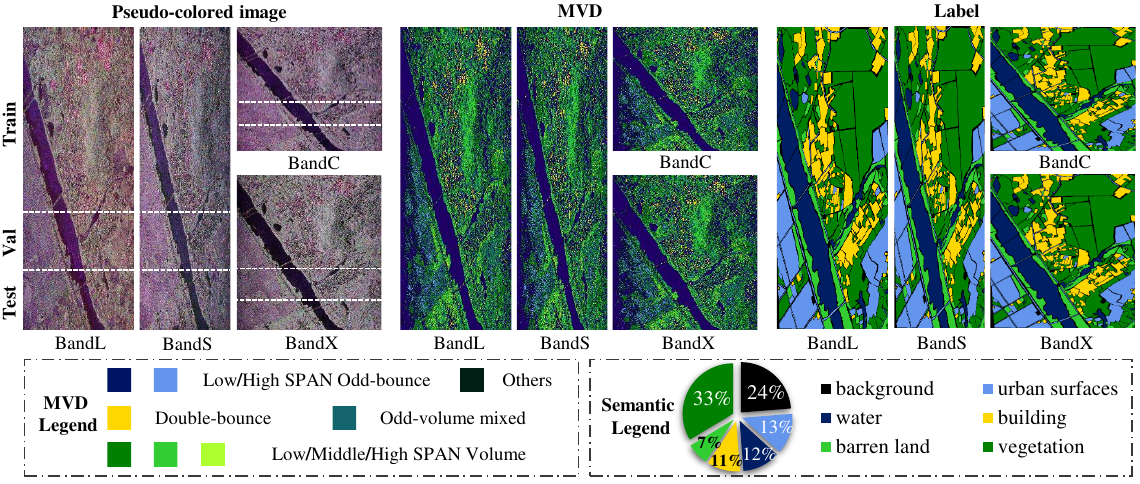}
    \caption{Visualization of the PhySAR-Seg-1 dataset, including the pseudo-colored image, MVD, semantic label, MVD legend, and semantic legend. The white dashed lines on the pseudo-colored image divide the dataset into train, val, and test sets.} 
    \label{fig::data_1}
\end{figure*}


\begin{figure*}[!htbp] 
    \centering 
    \includegraphics[width=1\textwidth]{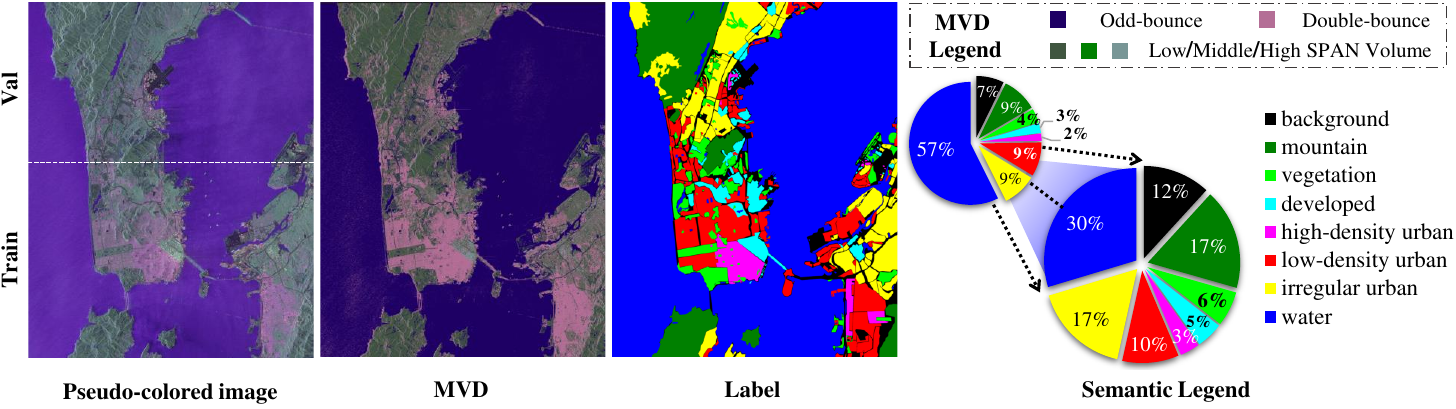}
    \caption{Visualization of the PhySAR-Seg-2 dataset, including the pseudo-colored image, MVD, semantic label, MVD legend, and semantic legend. The white dashed lines on the pseudo-colored image divide the dataset into train and val sets.} 
    \label{fig::data_SF}
\end{figure*}
\subsection{PhySAR-Seg Dataset Processing Pipeline}
The dataset processing pipeline involves three key steps: (1) Polarimetric decomposition and unsupervised GD-Wishart classification of PolSAR images; (2) MVD generation by re-clustering the physical scattering mechanisms based on semantic correlations; and (3) Manual annotation. Fig. \ref{fig::data_process} illustrates the entire process, using the PhySAR-Seg-1 dataset as an example.

In the first step, the pseudo-colored image is generated from the original single look complex (SLC) data through the Pauli decomposition algorithm, where three components are extracted and assigned to the RGB channels, respectively \cite{9169671}. Then, the SLC data is processed using the GD-Wishart classification algorithm \cite{ratha2017unsupervised}, yielding initial clustering results. The targets are categorized into three classes: odd-bounce, double-bounce, and volume. Notably, in practical experiments, double-bounce scattering is relatively rare, represented by only one class, while the other categories are further divided into five sub-classes each. This fine-grained sub-classification, combined with statistical cluster merging, inherently suppresses localized pixel-level noise from the unsupervised stage.

To better align MVD with semantic information, we re-cluster scattering classes by grouping sub-classes within each primary scattering category based on semantic correlations. Specifically, we compare these physical clusters with the Pauli pseudo-colored images to manually merge misaligned sub-classes, effectively correcting unsupervised classification errors. In addition to the three primary scattering types, the dataset includes mixed and 'other' classes, where the 'other' class typically contains boundary pixels with undefined scattering characteristics. After color-coding and visualizing the clusters, we generate MVD, a compact and informative representation of scattering mechanisms. MVD supports efficient large-scale processing, provides intuitive visualizations of electromagnetic properties, and preserves physical meaning, optimizing storage and enhancing interpretability for further analysis. Consequently, this generation pipeline is robustly applicable across diverse scenes and sensor platforms, yielding semantically consistent MVD products for each band, while the specific mechanism assignments naturally vary with wavelength, as quantified in the supplementary material.

\begin{figure*}[!htbp]
    \centering
    \includegraphics[width=0.99\textwidth]{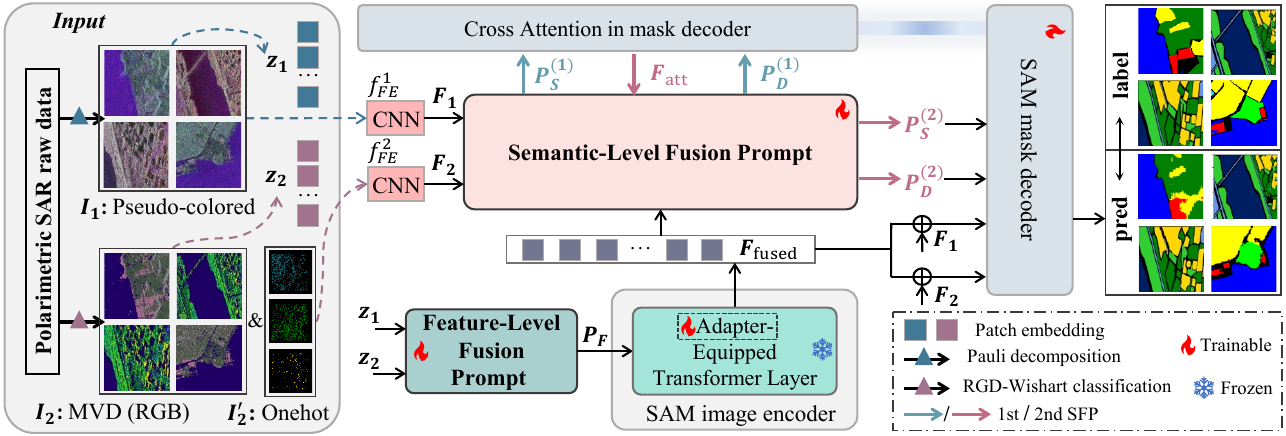}
    \caption{Overview of the proposed PolSAM. It consists of a novel data product, MVD, a Feature-Level Fusion Prompt (FFP) module and a progressive Semantic-Level Fusion Prompt (SFP) module. MVD is lightweight, encapsulates physical information, and correlates with semantic information. The FFP module combines data in a trainable manner for initial integration, while the SFP module generates sparse and dense prompts through a progressive two-level information interaction, guiding the segmentation task. By leveraging domain-specific knowledge of PolSAR data, PolSAM achieves enhanced segmentation performance with high data efficiency.} 
    \label{fig::model}
\end{figure*}

\subsection{Datasets Description}
Based on the above processing pipeline, we constructed two lightweight, physically interpretable, and information-integrated PhySAR-Seg datasets, each consisting of paired pseudo-colored and MVD images. Their comprehensive representation capabilities provide a solid foundation for evaluating our proposed method.

The PhySAR-Seg-1 dataset is sourced from \cite{yan2024mpolsar}. For our experiments, we use the L, S, C, and X bands, all covering the same geographic region. The L, S, and C bands have a resolution of 0.5 m, while the X band has a resolution of 1.0 m. Fig. \ref{fig::data_1} provides an overview of the dataset, showcasing the pseudo-colored images, MVD, and label maps, along with both the MVD and semantic legends. In the MVD legend, colors within the same hue represent consistent scattering types, with lighter shades corresponding to higher SPAN values. The semantic legend displays the proportions of the five final terrain classes, clearly illustrating their distribution across the dataset. Initially, the dataset contains 14 land cover types, which we merge into five categories: urban surfaces, water, buildings, barren land, and vegetation, as shown in the appendix. A brief note is that we calculate the MVD for each of the four bands. Upon comparison, the X band shows the best performance. Physically, the shorter X-band wavelength enhances sensitivity to small-scale roughness and fine structures, yielding more distinct and semantically separable scattering classifications. Since all bands cover the same region and capture similar scattering mechanisms, we register the X-band MVD to the other three bands to generate the final result. The segmented dataset comprises 2,866 image pairs of pseudo-colored MVD, each with 512 $\times$ 512 pixels. We divide the data from each band into training, validation, and test sets in a 6:2:2 ratio, following the dashed lines shown in Fig. \ref{fig::data_1}. This approach ensures that data from the same geographic region in different bands is allocated to the same set, preventing any overlap between the training, validation, and test sets.

The PhySAR-Seg-2 dataset is sourced from \cite{Gaofen3} and is based on spaceborne C-band data. We also obtain MVD according to the method in Fig. \ref{fig::data_process}, and labels are obtained through manual annotation, resulting in a total of seven categories excluding the background. Fig. \ref{fig::data_SF} provides an overview of the dataset, including pseudo-colored images, MVD, and their corresponding semantic labels. Due to the limited number of samples in the dataset, we only partition the data into a training set and a validation set in a 6:4 ratio, as indicated by the dashed line in the figure. Given the substantial presence of water in the dataset, we strategically remove pure water samples from the segmented 512 $\times$ 512 images to maintain a balance among categories. After this adjustment, 1,110 images remain, with 668 allocated to the training set and 442 to the validation set. The semantic legend clearly indicates the category distribution before and after processing.


\section{The Proposed Method}
\label{sec:method}
\subsection{Revisiting SAM}
The SAM introduces an innovative prompt-based architecture for diverse segmentation tasks. It consists of three main components: the image encoder ($f_\text{Enc}$), the prompt encoder ($f_\text{PE}$), and the mask decoder ($f_\text{Dec}$). The image encoder extracts high-dimensional feature maps from the input image, which serve as the foundation of SAM’s segmentation capability. The prompt encoder plays a key role by accepting flexible inputs, such as sparse prompts ($I_{\text{sparse}}$, e.g., points or bounding boxes), dense prompts ($I_{\text{dense}}$, e.g., masks), and text. These prompts guide the model to focus on specific regions of interest and are transformed into embeddings that integrate with the image features. Finally, the mask decoder fuses both the sparse and dense embeddings with the image features to produce the final segmentation mask. This approach enables SAM to handle diverse segmentation challenges effectively without task-specific fine-tuning. It is a versatile and robust solution for a wide range of scenarios. This can be mathematically represented as follows:
\begin{equation}
\begin{aligned}
    \mathbf{F}_{\text{img}} &= f_\text{Enc}(I), \\
    \mathbf{P}_{\text{s}}  , \mathbf{P}_{\text{d}} &= f_\text{PE}(I_{\text{sparse}}, I_{\text{dense}}), \\
    \mathbf{M} &= f_\text{Dec}(\mathbf{F}_{\text{img}}, \mathbf{P}_{\text{s}}, \mathbf{P}_{\text{d}}),
\end{aligned}
\end{equation}
where $I$ represents the input image, $\mathbf{F}_{\text{img}}$ is the image feature map extracted by the image encoder. $\mathbf{P}_{\text{s}}$, $\mathbf{P}_{\text{d}}$ represent the sparse prompt embeddings and dense prompt embeddings. $\mathbf{M}$ is the final segmentation mask produced by the mask decoder.

Building on the success of SAM in the segmentation field, we adopt its image encoder and mask decoder as the backbone of our segmentation network, keeping the encoder frozen. Inspired by SAM's prompt-based mechanism, which relies on geometric prompts such as points, boxes, or masks, we propose a novel approach that autonomously generates semantic prompts. By incorporating domain-specific cues, our method produces both sparse and dense semantic prompts, enabling the model to effectively leverage task-specific and data-specific information for enhanced segmentation performance. The detailed structure is described below.
\subsection{PolSAM Overview}

The overview of our proposed PolSAM is illustrated in Fig. \ref{fig::model}. We adopt the SAM image encoder ($f_\text{Enc}$) and mask decoder ($f_\text{Dec}$), keeping their original structures unchanged. Pseudo-colored images offer rich texture details and strong visual representation, while MVD captures scattering mechanisms and conveys semantic information. To effectively exploit and integrate these data-specific features, we propose the Feature-Level Fusion Prompt (FFP) module, placed before the image encoder to fuse features from both inputs for a more comprehensive representation. Additionally, to address data modality incompatibility and better adapt the encoder to PolSAR data, we introduce trainable adapter modules into each layer of the frozen image encoder, enhancing feature representation and segmentation performance.

To provide effective semantic prompts to the decoder, we designed a progressive Semantic-Level Fusion Prompt (SFP) module. Each level targets different aspects of key information, enabling nuanced semantic integration. The process is described by the following equations:

\begin{equation}
\begin{aligned}
    \mathbf{F}_{\text{fused}} &= f_\text{Enc} (\textcolor{purple}{f_\text{FFP}}
    (I_{\text{1}}, \textcolor{blue}{I_{\text{2}}})), \\
    \mathbf{P}_{\text{s}}  , \mathbf{P}_{\text{d}} &= \textcolor{purple}{f_\text{SFP}}(I_{\text{1}}, \textcolor{blue}{I'_{\text{2}}}), \\
    \mathbf{M} &= f_\text{Dec}(\mathbf{F}_{\text{fused}}, \mathbf{P}_{\text{s}}, \mathbf{P}_{\text{d}}),
\end{aligned}
\label{eq2}
\end{equation}
where, $f_\text{FFP}$ and $f_\text{SFP}$ represent the FFP and SFP modules, respectively.  $I_\text{1}$ denote the pseudo-colored image, $I_\text{2}$ and $I'_\text{2}$ represent MVD in RGB and one-hot form respectively. In the following sections, we will analyze each module in details.
\subsection{Feature-Level Fusion Prompt}

The design of FFP follows the concept proposed in \cite{jia2022visual}, where a lightweight module with convolutions is inserted into the beginning of the image encoder. In this context, the pseudo-colored image and MVD image are denoted as $I_1$ and $I_2$, respectively, with their corresponding patch embeddings represented by $\mathbf{z}_1$ and $\mathbf{z}_2$. Given the distinct characteristics of these two inputs, $\mathbf{z}_1$ is closely related to local details and texture in the image, while $\mathbf{z}_2$ reflects the scattering mechanism, which is more related to semantic information. Therefore, enhancing the spatial resolution of $\mathbf{z}_1$ is essential to emphasize local details, whereas $\mathbf{z}_2$ undergoes direct dimensionality reduction to retain its global semantic features.

\begin{figure}[!htbp] 
    \centering 
    \includegraphics[width=0.47\textwidth]{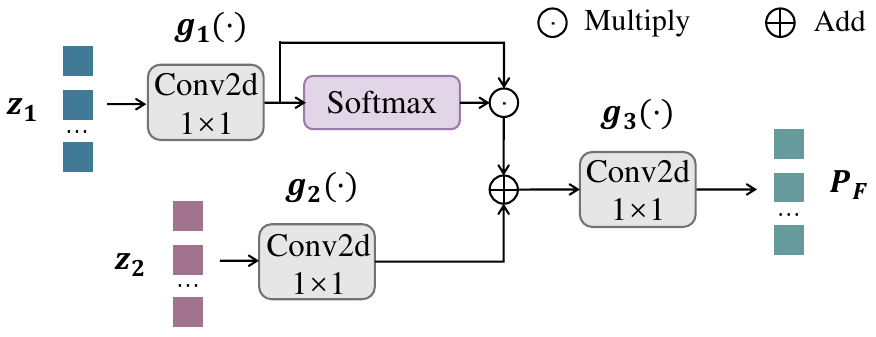}
    \caption{Detailed design of module FFP.} 
    \label{fig::detail1}
\end{figure}

Specifically, as shown in Fig. \ref{fig::detail1}, FFP first takes $\mathbf{z}_1$ and $\mathbf{z}_2$ as input and performs dimension reduction using 1$\times$1 convolutions to obtain the latent features for each input. We first apply a spatial Softmax activation to the latent features $\mathbf{z}_1$ to highlight key regions. Then, we use element-wise multiplication to enhance these areas. Subsequently, the fusion features are projected back to their original dimension using 1$\times$1 convolutions, yielding the initial fusion prompt at the feature level. Here, we use $g_1(\cdot)$, $g_2(\cdot)$, and $g_3(\cdot)$ to represent the three convolution layers. Therefore, the fusion prompt can be derived as:

\begin{equation}
\begin{aligned}
    \mathbf{P}_\mathrm{F} &= f_\text{FFP}(\mathbf{z}_1,\mathbf{z}_2) \\
    & = g_3(g_1(\mathbf{z}_1)  \odot \text{Softmax}(g_1(\mathbf{z}_1)) + g_2(\mathbf{z}_2)).
\end{aligned}
\end{equation}

\subsection{Adapter-Equipped Image Encoder}

We add $\mathbf{P}_\mathrm{F}$ to $\mathbf{z}_1$ and $\mathbf{z}_2$ before feeding them into the frozen pre-trained image encoder. To adapt feature extraction for PolSAR data under a supervised fine-tuning paradigm, we insert trainable adapters \cite{wu2025medical} into each encoder layer. In this way, the fusion prompt and the adapters are learned to satisfy the complementarity of different inputs by fine-tuning only a few parameters. Thus, we can detail the acquisition of $\mathbf{F}_{\text{fused}}$ as presented in Equation \eqref{eq2}:
\begin{equation}
    \mathbf{F}_{\mathrm{fused}} = f_\text{Enc}(\mathbf{P}_\mathrm{F},\mathbf{z}_1,\mathbf{z}_2).
    \label{con:Ffused}
\end{equation}

\begin{figure}[!htbp] 
    \centering 
    \includegraphics[width=0.47\textwidth]{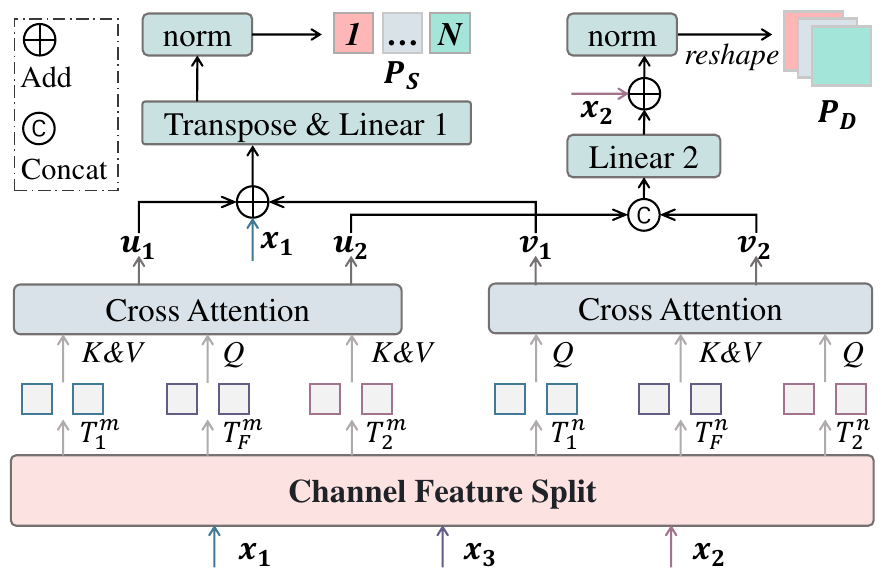}
    \caption{Detailed design of the proposed SFP: $\mathbf{P}_\mathrm{S}\in \mathbb{R}^{N \times C}$ and $\mathbf{P}_\mathrm{D} \in \mathbb{R}^{H \times W \times C}$ represent sparse prompts and dense prompts, respectively.} 
    \label{fig::detail2}
\end{figure}
\subsection{Semantic-Level Fusion Prompt}

Inspired by SAM's promptable mechanism, we propose the SFP module, which generates sparse and dense prompt embeddings to guide segmentation. Featuring a two-level progressive structure, the SFP module optimizes feature integration and enhances semantic representation. This design balances the extraction of unique and complementary information from different inputs, enriching the segmentation process. By refining feature interactions progressively, the SFP module significantly improves segmentation accuracy and contextual awareness.

First, the pseudo-colored images $I_1$ are processed using a basic convolutional block and two pooling layers to obtain the feature embedding $\mathbf{F}_1$, denoted as $f_\text{FE}^1(\cdot)$, as shown in Fig. \ref{fig::model}. Considering the strong correlation between MVD and semantic labels, for MVD, we directly adopt the CNN network from the SAM prompt encoder used for processing mask prompts. This layer downsamples the input using three sequential convolutional layers. Intermediate normalization and activation functions reduce the channel dimension and map the output to the target embedding space. To ensure that the semantic prompts are not affected by color encoding, we directly use the one-hot form of the MVD, denoted as $I'_2$. We then obtain the feature embedding $\mathbf{F}_2$, represented by $f_\text{FE}^2(\cdot)$. The features $\mathbf{F}_1$ and $\mathbf{F}_2$ have the same dimension as $\mathbf{F}_{\text{fused}}$. The above can be denoted as:

\begin{equation}
    \mathbf{F}_1 = f_\text{FE}^1(I_1),  \quad \quad   \mathbf{F}_2 = f_\text{FE}^2(I'_2).
\end{equation}

The proposed SFP is a progressive process, with two levels implemented using a shared module. To clarify, we denote them as SFP-1 and SFP-2. As shown in Fig. \ref{fig::model}, each level of the SFP module receives three inputs, denoted as $f_\text{SFP}(x_1, x_2, x_3)$. The inputs of the SFP-1 are the single input embedding $\mathbf{F}_1$ and $\mathbf{F}_2$, together with the fused feature from encoder $\mathbf{F}_{\mathrm{fused}}$. It outputs the initial sparse and dense prompt $\mathbf{P}_\mathrm{S,D}^{(1)}$. Then, to make the semantic prompts more specific, SFP-2 combines the single-input features and the fusion features in an additive manner, and interacts with the semantic features $\mathbf{F}_{\mathrm{att}}$ after the mask decoder attention layer to generate the final sparse and dense prompt $\mathbf{P}_\mathrm{S,D}^{(2)}$, $\mathbf{P}_\mathrm{S}\in \mathbb{R}^{N \times C}$ and $\mathbf{P}_\mathrm{D} \in \mathbb{R}^{H \times W \times C}$. The process can be denoted as:
\begin{equation}
\begin{aligned}
    \mathbf{P}_\mathrm{S/D}^{(1)} &= f_\text{SFP}(\mathbf{F}_\mathrm{1}, \mathbf{F}_\mathrm{2}, \mathbf{F}_{\mathrm{fused}}), \\
    \mathbf{P}_\mathrm{S/D}^{(2)} &= f_\text{SFP}(\mathbf{F}_{\mathrm{fused}} + \mathbf{F}_\mathrm{1}, \mathbf{F}_{\mathrm{fused}} + \mathbf{F}_\mathrm{2}, \mathbf{F}_{\mathrm{att}}),
\end{aligned}
\end{equation}

The details of SFP is shown in Fig. \ref{fig::detail2}. Given three inputs $x_1, x_2, x_3 \in \mathbb{R}^{C \times H \times W}$, the ``Channel Feature Split" expands their channel dimension from $C$ to $2C$ via a linear projection and ReLU activation. These expanded features are then evenly halved along the channel dimension to yield two independent groups of $\mathbb{R}^{C \times H \times W}$, denoted as $[T_1^m, T_2^m, T_3^m]$ and $[T_1^n, T_2^n, T_3^n]$, respectively. By exchanging the query and value, two groups of enhanced features can be obtained with cross-attention modules, denoted as $[v_1,v_2]$ and $[u_1,u_2]$, with the dimension of $HW \times C$. Then, the sparse prompt and dense prompt are generated separately with the following operations:

\begin{equation}
\begin{aligned}
    \mathbf{P}_\mathrm{S} & = \text{Norm}(\text{Linear}_1(0.5(v_{1}+u_{1})+ \text{DimTrans}(x_1))) \\
    \mathbf{P}_\mathrm{D} & = \text{Norm}(\text{Linear}_2(\text{Concat}(v_2,u_2))+ \text{DimTrans}(x_2)), \label{conv:P_SD12}
\end{aligned}
\end{equation}
where $\text{Norm}(\cdot)$, $\text{Concat}(\cdot)$, and $\text{DimTrans}(\cdot)$ denote the normalization, concatenation, and dimension transformation, respectively. The linear mappings $\text{Linear}_1(\cdot)$ and $\text{Linear}_2(\cdot)$ aim to map the dimension from $HW$ to $N$, from $2C$ to $C$, respectively.

From the above descriptions, we can find that SFP module aims to fuse different inputs interactively so that the information is enriched to a certain extent. The two cross-attention modules consider different inputs as the query to obtain the attention weight for feature enhancement. In the SFP-1, $\mathbf{F}_1$, $\mathbf{F}_2$, and $\mathbf{F}_\mathrm{fused}$ are corresponding to $x_1$, $x_2$, and $x_3$. Specifically, $u_1$ and $u_2$ represent the segmentation fusion features ($\mathbf{F}_\mathrm{fused}$) enhanced by the single-input features. Conversely, $v_1$ and $v_2$ denote the single-input features refined by the fusion information. The initial sparse and dense semantic prompts pass through the cross-attention module in the mask decoder to output $\mathbf{F}_\mathrm{att}$, which represents high-level semantic features. To this end, in the SFP-2, $u_1$, $u_2$ and $v_1$, $v_2$ indicates the result of the mutual enhancement between the fusion information and high-level semantic features.

The SFP module enhances feature representation by interactively fusing inputs and progressively refining semantic prompts. This ensures critical details are preserved while enriching the feature space. Additionally, the mutual enhancement between fused information and high-level semantic features improves the model's understanding, leading to better segmentation results.

\begin{table*}
    \centering
    \caption{Comparison of PolSAM performance on the PhySAR-Seg-1 dataset: methods above the dashed line are SAM-based, and those below are multimodal. Metrics include per-category IoU and overall performance. Best results are in bold, second-best underlined, and complete PolSAM is marked in red. For BER, a lower value indicates better performance.}
    \label{Data1}
    \resizebox{0.98\textwidth}{!}{
    \renewcommand{\arraystretch}{0.68}
    \begin{tabular}{ccccccccccc}
      \toprule
      \multirow{2}{*}{\textbf{Models}} &\multicolumn{6}{c}{\textbf{IoU Per Category (\%)}} &\multicolumn{4}{c}{\textbf{Overall Metrics (\%)}}  \\
     \cmidrule(r){2-7}
     \cmidrule(r){8-11}
     &BK &Urban surfaces &Water &Building &Barren land &Vegetation  &mAcc &mF1 score &mIoU &BER  \\
    \midrule
    HQ-SAM\cite{ke2024segment}     &22.05 &4.92 &10.76 &1.05 &1.01 &27.34  &31.49 &25.70 &11.19 &71.76 \\
    SAM-LST\cite{chai2024ladder}   &32.26 &60.20 &79.60 &57.80 &54.73 &66.14  &72.72 &72.70  &58.45 &29.45 \\
    Personalize-SAM\cite{zhang2024personalize}  &44.65 &65.12 &83.99 &60.93 &73.29 &69.8   &78.26 &79.39 &66.30 &23.97  \\
    Mobile-SAM\cite{zhang2023faster}  &22.91 &52.09 &77.28 &42.85 &44.24 &55.62  &64.18 &64.84  &49.17 &38.43 \\
    RSAM-Seg\cite{zhang2025rsam} &\underline{49.55} &\underline{69.32} &\underline{85.63} &\underline{66.15} &\underline{74.60} &\underline{73.23}  &\underline{81.18} &\underline{81.67}  &\underline{69.75} &\underline{18.43} \\
    \rowcolor{gray!20}  \textbf{PolSAM(w/o MVD)} &\textbf{51.22} &\textbf{69.64} &\textbf{86.45} &\textbf{68.98} &\textbf{78.52} &\textbf{76.21}   & \textbf{82.62} & \textbf{83.37} &\textbf{71.84} &\textbf{16.33}     \\
    \hdashline
    MMSFormer\cite{reza2024mmsformer} &46.19 &68.79 &85.14 &63.35 &67.57 &72.64  &79.52 &80.03 &67.28 &21.17 \\
    SFAF-MA\cite{he2023sfaf}    &\underline{54.29} &\underline{73.03} &\underline{87.56} &\underline{69.41} &\underline{77.07} &\textbf{76.99}   &\underline{83.67} &\underline{84.33} &\underline{73.06} &\underline{15.89} \\
    GMNet\cite{zhou2021gmnet}   &45.41 &65.11 &50.03 &53.05 &65.16 &70.74 &75.96 &75.62 &58.25 &37.43  \\
    CMX\cite{zhang2023cmx}  &36.52 &65.07 &46.87 &63.58 &64.5 &60.18  &69.15 &73.24 &56.12 &29.79\\
    LSNet\cite{zhou2023lsnet} &28.87 &59.83 &79.45 &53.69 &47.04 &57.07  &67.55 &72.76  &54.49 &32.07 \\
    \rowcolor{gray!38} \textbf{PolSAM(w/ MVD)}  &\textbf{55.44} &\textbf{73.97} &\textbf{87.97} &\textbf{69.62} &\textbf{78.66} &\underline{75.65}  & \textbf{\textcolor{red}{83.70}} & \textbf{\textcolor{red}{84.71}}  &\textbf{\textcolor{red}{73.55}} &\textbf{\textcolor{red}{14.75}} \\
    \bottomrule  \end{tabular}
}
\end{table*}

\begin{table*}
    \centering
    \caption{Comparison of PolSAM performance on the PhySAR-Seg-2 dataset. For BER, a lower value indicates better performance.}
    \label{Data2}
    \resizebox{\textwidth}{!}{
    \renewcommand{\arraystretch}{0.68}
    \begin{tabular}{ccccccccccccc}
      \toprule
        \multirow{2}{*}{\textbf{Models}} &\multicolumn{8}{c}{\textbf{IoU Per Category (\%)}} &\multicolumn{4}{c}{\textbf{Overall Metrics (\%)}}  \\
     \cmidrule(r){2-9}
     \cmidrule(r){10-13}
     &BK &Montain &Vegetation &Developed &I-urban &H-urban &L-urban &Water  &mAcc &mF1 score &mIoU &BER  \\
    \midrule
    HQ-SAM\cite{ke2024segment}     &7.91 &15.89 &0.77 &0.11 &0.00 &6.88 &7.98 &44.71   &29.69 &21.88 &10.53 &78.83  \\
    SAM-LST\cite{chai2024ladder}   &\underline{29.22} &\textbf{78.26} &\underline{18.27} &\underline{15.55} &\underline{27.44} &32.67 &47.75 &93.38 &\underline{74.19} &\underline{58.29} &\underline{42.82} &\underline{44.38} \\
    Personalize-SAM\cite{zhang2024personalize}    &28.61 &65.63 &14.68 &5.74 &1.74 &32.06 &47.06 &\textbf{94.62}  &70.24 &50.39 &36.27 &51.83 \\
    Mobile-SAM\cite{zhang2023faster}    &28.86 &68.54 &13.18 &5.91 &1.98 &\underline{33.98} &\underline{50.74} &93.87  &70.76 &49.07 &37.13 &48.69 \\
    RSAM-Seg\cite{zhang2025rsam}    &28.11 &73.45 &12.47 &11.24 &17.99 &33.32 &35.64 &\underline{94.58}  &68.59 &56.38 &38.35 &48.65 \\
    \rowcolor{gray!20} \textbf{PolSAM(w/o MVD)}    &\textbf{35.00} &\underline{77.15} &\textbf{21.07} &\textbf{16.27} &\textbf{34.16} &\textbf{35.08} &\textbf{50.91} &92.76  &\textbf{74.31}  & \textbf{59.03} & \textbf{45.30} &\textbf{40.62}  \\
    \hdashline
    MMSFormer\cite{reza2024mmsformer}  &30.84 &72.00 &14.28 &8.60 &0.00 &\underline{37.93} &44.60 &93.80  &72.24 &51.51 &37.75 &50.75 \\
    SFAF-MA\cite{he2023sfaf}           &\underline{32.78} &\underline{81.39} &17.74 &12.39 &5.82 &33.28 &52.62 &94.54  &74.49 &52.69 &41.32 &45.98 \\
    GMNet\cite{zhou2021gmnet}   &31.75 &\textbf{82.73} &\textbf{24.75} &16.17 &11.08 &36.39 &\textbf{55.23} &\textbf{95.22}  &\textbf{77.31} &\underline{60.92} &\underline{44.17} &\underline{41.19} \\
    CMX\cite{zhang2023cmx}    &29.21 &73.12 &14.20 &\textbf{22.02} &2.14 &36.93 &41.32 &\underline{94.90}   &69.75 &53.84 &39.23 &46.17\\
    LSNet\cite{zhou2023lsnet}   &27.70 &75.36 &18.31 &0.12 &\underline{19.93} &29.06 &46.14 &94.06   &70.68 &53.11 &38.83 &45.83 \\
    \rowcolor{gray!38} \textbf{PolSAM(w/ MVD)}    &\textbf{34.83} &75.88 &\underline{22.63} &\underline{18.13} &\textbf{30.76} &\textbf{42.43} &\underline{54.57} &93.31   &\underline{\textcolor{red}{76.47}}  & \textbf{\textcolor{red}{61.40}} & \textbf{\textcolor{red}{46.55}} &\textbf{\textcolor{red}{39.07}}    \\
    \bottomrule  \end{tabular}
}
\end{table*}

\subsection{Loss Function}
The feature embedding of single and fused inputs are then added together as the input of mask decoder, as well as the sparse and dense semantic prompt. Finally, the segmentation result $y$ can be obtained as:
\begin{equation}
    y = f_\text{Dec}(\mathbf{F}_\mathrm{1}, \mathbf{F}_\mathrm{2},\mathbf{F}_{\mathrm{fused}},\mathbf{P}_\mathrm{S/D}). \label{con:Dec}
\end{equation}

Based on extensive experimental comparisons, we use the cross-entropy loss function for the PhySAR-Seg-1 dataset, which is defined as:
\begin{equation}
    \mathcal{L}_{\text{CE}} = -\sum_{i=1}^{N} y_i \log(\hat{y}_i),
\end{equation}
where \(y_i\), \(\hat{y}_i\) are the true label and predicted probability for class \(i\), and \(N\) is the number of samples.

For PhySAR-Seg-2 dataset, where the class distribution is more unbalanced, we adopt the focal loss function to calculate the loss. The weight for each category is the inverse of its proportion in the dataset, enhancing the model's focus on harder-to-classify, less frequent categories. The focal loss is given by:
\begin{equation}
    \mathcal{L}_{\text{FL}} = -\sum_{i=1}^{N} w_i (1 - \hat{y}_i)^\gamma y_i \log(\hat{y}_i),
\end{equation}
where \(w_i = \frac{1}{p_i}\) is the weight for class \(i\) (with \(p_i\) being the proportion of class \(i\)), \(\gamma\) is the focusing parameter to down-weight easy examples, \(y_i\) is the true label, and \(\hat{y}_i\) is the predicted probability for class \(i\).
\section{Experiments and Discussions}
\label{sec:exp}
This section presents experimental configuration, results analysis, and discussions, including metric comparisons and visualization interpretations. It encompasses ablation studies on module design, evaluation of semantic prompt effectiveness, hyperparameter analysis of sparse prompts, and discussion on MVD effectiveness.

\subsection{Experimental Settings}

\noindent \textbf{Implementation Details.}The pre-trained ViT-B model in SAM serves as our encoder backbone. The framework is trained using the AdamW optimizer with $\beta_1 = 0.9$ and $\beta_2 = 0.999$. We set the batch size to 12, epochs to 350, and the initial learning rate to 0.001. A 4\% training warm-up is applied, followed by exponential decay of the learning rate with a factor of 0.9. Experiments are conducted on a single NVIDIA 4090Ti GPU.

\noindent \textbf{Evaluation Metrics}. Following existing methods, we apply three common overall metrics to evaluate the results of the semantic segmentation, including mean pixel accuracy (mAcc), mean F1 score (mF1 score), and mean intersection over union (mIoU). Additionally, for each class, we use the IoU as the evaluation metric. Furthermore, considering the class imbalance of the datasets, we report the
Balanced Error Rate (BER), defined as the average of per-class error rates:
$\mathrm{BER} = \frac{1}{K}\sum_{i=1}^{K}\left(1-\frac{\mathrm{TP}_i}
{\mathrm{TP}_i+\mathrm{FN}_i}\right)$, where $K$ is the number of classes, and
$\mathrm{TP}_i$ and $\mathrm{FN}_i$ denote the true positive and false negative pixels of
class $i$. A lower BER indicates more balanced performance across categories.

\begin{figure*}[htbp]
\centering
\includegraphics[width=\textwidth]{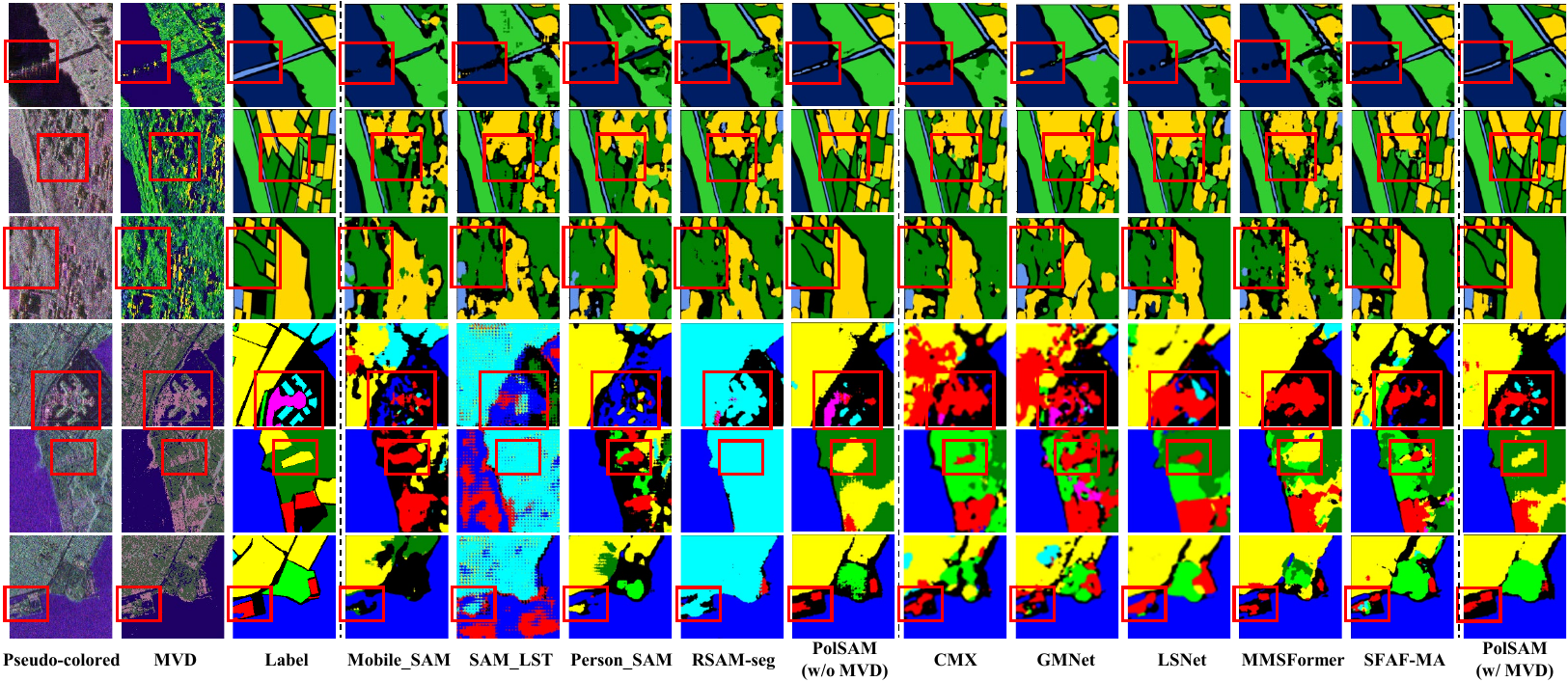}
\caption{Visualization of segmentation results on the PhySAR-Seg-1 (rows 1-3) and PhySAR-Seg-2 (rows 4-6) datasets: each row displays a randomly selected image from the test set, while each column shows results from different models.}
\label{fig:visual}
\end{figure*}

\begin{figure*}[!htbp]
    \centering 
    \includegraphics[width=1\textwidth]{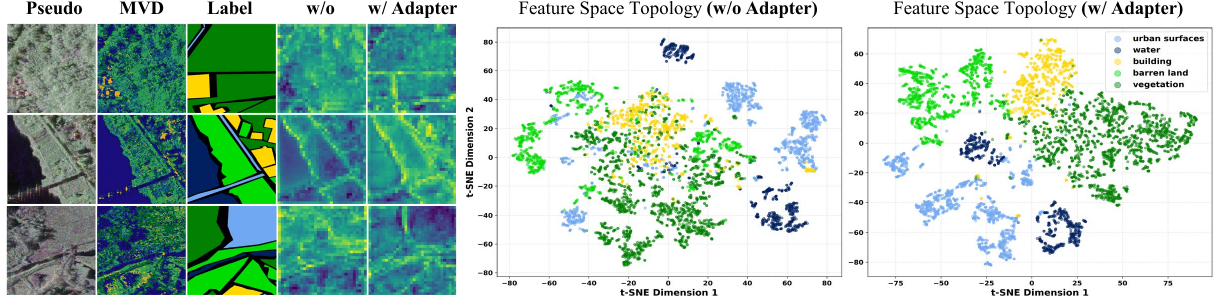}
    \caption{Visualization of feature activation maps and t-SNE distributions before (w/o) and after (w/) applying adapters. The adapters significantly enhance feature discriminability and semantic clustering, mitigating the modality gap.}
    \label{fig::adapter}
\end{figure*}

\subsection{Comparison With State-of-the-art Methods}
Our comprehensive evaluation benchmarks PolSAM against ten SOTA methods. Given the original SAM's single-input constraint, we first compare the pseudo-color-only variant (PolSAM w/o MVD) with five existing SAM-based approaches \cite{ke2024segment,chai2024ladder,zhang2023faster,zhang2024personalize,zhang2025rsam}. Subsequently, to demonstrate the advantage of our dual-input design, we compare the full PolSAM with five multimodal fusion algorithms \cite{reza2024mmsformer,he2023sfaf,zhou2021gmnet,zhang2023cmx,zhou2023lsnet}. The experimental results are detailed in Tables \ref{Data1} and \ref{Data2} and discussed below.

\noindent \textbf{Evaluation Metrics.}
As shown in Table \ref{Data1}, on the PhySAR-Seg-1 dataset, our single-input variant, PolSAM (w/o MVD), surpasses the top-performing SAM-based method, RSAM-Seg \cite{zhang2025rsam}, with improvements of 1.44\% in mAcc, 1.7\% in mF1, and 2.09\% in mIoU. Similarly, on PhySAR-Seg-2 (Table \ref{Data2}), it outperforms the leading competitor, SAM-LST \cite{chai2024ladder}, by 2.48\% in mIoU. Furthermore, the complete PolSAM framework achieves state-of-the-art performance against multimodal fusion methods, exceeding GMNet \cite{zhou2021gmnet} by 2.38\% in mIoU on the second dataset. In terms of BER, PolSAM (w/ MVD) attains the lowest values on both datasets
(14.75\% and 39.07\%). Notably, on PhySAR-Seg-2 it yields a lower BER than GMNet despite
the slightly lower mAcc, indicating more balanced predictions over the minority urban
categories.

Overall, the results highlight three key strengths: first, PolSAM demonstrates superior generalization, outperforming competitors across most categories; second, the integration of MVD consistently boosts performance over the single-input baseline; and third, the model proves particularly robust in challenging scenarios—such as the 'urban' class in PhySAR-Seg-2, where it effectively captures double-bounce scattering despite limited sample proportions.

\noindent \textbf{Visualization.}
Figure \ref{fig:visual} presents a qualitative comparison between SAM-based models and multimodal fusion methods on the PhySAR-Seg-1 (rows 1-3) and PhySAR-Seg-2 (rows 4-6) datasets. PolSAM achieves the most visually coherent results by effectively combining the strengths of its inputs: the pseudo-colored images provide fine-grained texture, while the MVD offers clear high-level semantic distinctions. This effective fusion of feature and semantic levels enables the model to generate precise prompts that guide the decoder for optimal segmentation.

As highlighted by the red boxes, PolSAM excels at preserving boundary details. While methods like LSNet \cite{zhou2023lsnet} incorporate edge constraints to enhance performance, they are still limited by the intrinsic speckle noise and ambiguous edges of PolSAR imagery. PolSAM overcomes this by prioritizing semantic consistency, yielding cleaner boundaries and superior segmentation performance compared to edge-constrained baselines. Nevertheless, PolSAM occasionally yields fragmented edge predictions in highly ambiguous regions, such as the urban transition zones in PhySAR-Seg-2 (Figure \ref{fig:visual}, rows 4 and 5). These failures mainly occur in two scenarios: severe scattering mixtures, where multiple scattering mechanisms coexist within a resolution cell and blur the class evidence; and low-contrast boundaries under intrinsic speckle noise, where adjacent categories exhibit similar polarimetric responses (e.g., high- and low-density urban areas). Potential remedies include boundary-aware constraints and uncertainty-guided prompt refinement, which we leave for future work.

\begin{table*}[htbp]
\caption{Ablation studies on module design of PolSAM on PhySAR-Seg-1 and PhySAR-Seg-2. For BER, a lower value indicates better performance.}
\label{AS1}
\renewcommand{\arraystretch}{0.68}
\centering
\resizebox{\textwidth}{!}{
\begin{tabular}{lcccccccccccccc}
    \toprule
  \rowcolor{gray!38} &\multicolumn{1}{c}{\textbf{Data}} &\multicolumn{2}{c}{\textbf{Encoder}}  &\multicolumn{3}{c}{\textbf{Decoder}}&\multicolumn{4}{c}{\textbf{PhySAR-Seg-1(\%)}}  &\multicolumn{4}{c}{\textbf{PhySAR-Seg-2(\%)}}\\
    \cmidrule(r){2-2} \cmidrule(r){3-4}   \cmidrule(r){5-7} \cmidrule(r){8-11} \cmidrule(r){12-15}
  \rowcolor{gray!20} \quad &MVD &Adapter  &FFP  & $PE_{1/2}$  &SFP-1  &SFP-2  &mAcc  &mF1 score  &mIoU &BER &mAcc  &mF1  score  &mIoU &BER\\
   \cmidrule(r){1-15}
    Baseline &\textcolor{blue}{\ding{55}}  &\textcolor{blue}{\ding{55}} &\textcolor{blue}{\ding{55}}  &\textcolor{blue}{\ding{55}} &\textcolor{blue}{\ding{55}}  &\textcolor{blue}{\ding{55}}   & 70.83  & 71.42 & 56.49 &28.73 & 63.48  & 43.29 & 32.05 &50.64 \\
    M1  &\usym{1F5F8} &\textcolor{blue}{\ding{55}}   &\textcolor{blue}{\ding{55}}   &\usym{1F5F8}   &\usym{1F5F8}  &\usym{1F5F8}     & 72.83  & 73.96 & 59.55 &25.43 & 70.77  & 55.38 & 41.00 &44.89\\
    M2      &\usym{1F5F8}   &\textcolor{blue}{\ding{55}}  &\usym{1F5F8}   &\usym{1F5F8}  &\usym{1F5F8}  &\usym{1F5F8}  & 75.41  & 76.51 & 62.76 &24.19 & 74.63  & 57.21 & 43.20 &43.56\\
    M3     &\usym{1F5F8}    &\usym{1F5F8}    &\textcolor{blue}{\ding{55}}   &\usym{1F5F8}  &\usym{1F5F8} &\usym{1F5F8}  & 82.37  & 83.44 & 71.33 &16.64 & 74.94  & 59.78 & 45.38 &40.73\\
    M4     &\usym{1F5F8}   &\usym{1F5F8}      &\usym{1F5F8}  &\textcolor{blue}{\ding{55}} &\textcolor{blue}{\ding{55}} &\textcolor{blue}{\ding{55}}     & 80.64  & 81.71 & 69.58 &17.69 & 72.26  & 53.38 & 40.62 &45.93 \\
    M5     &\usym{1F5F8}   & \usym{1F5F8}  &\usym{1F5F8}     &\usym{1F5F8} &\usym{1F5F8} &\textcolor{blue}{\ding{55}}  & 81.24  & 82.30 & 70.10 &17.86 & 73.17  & 57.77 & 41.61 &43.69\\
    PolSAM(w/o MVD)    &\textcolor{blue}{\ding{55}}    & \usym{1F5F8}  &\usym{1F5F8}    &\usym{1F5F8} &\usym{1F5F8} &\usym{1F5F8} & 82.62  & 83.37 & 71.84 &16.33 & 73.69  & 59.69 & 45.30 &40.62\\
    \textbf{PolSAM(w/ MVD)} &\usym{1F5F8} &\usym{1F5F8} &\usym{1F5F8}  &\usym{1F5F8} &\usym{1F5F8} &\usym{1F5F8} & \textbf{83.70}  & \textbf{84.71} & \textbf{73.55} &\textbf{14.75} & \textbf{76.47}  & \textbf{61.40} & \textbf{46.55} &\textbf{39.07}\\
    \bottomrule
\end{tabular}
}
\end{table*}

\begin{figure*}[htbp]
\centering
\includegraphics[width=\textwidth]{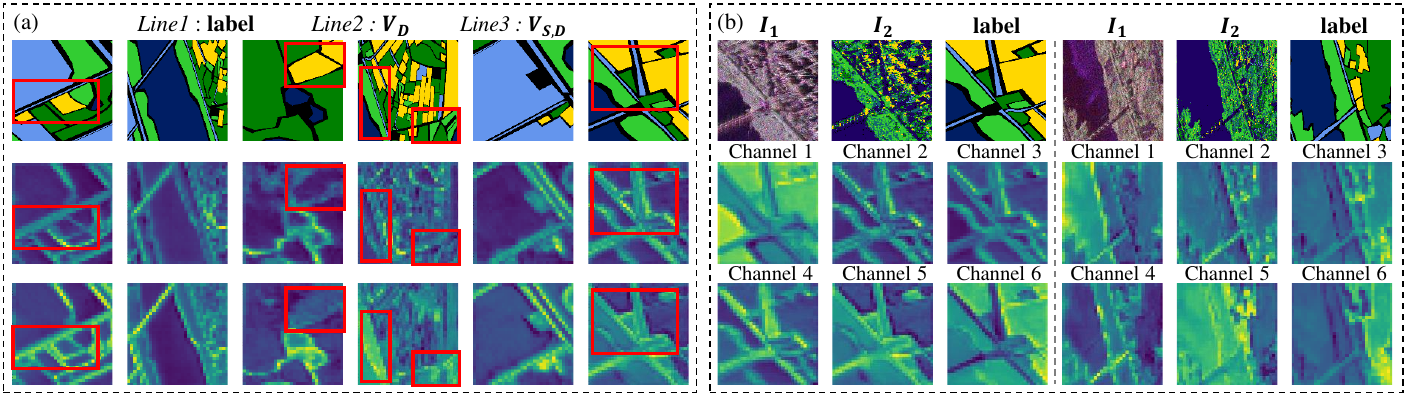}
\caption{(a) Visualization of labels, dense prompt embedding $\mathbf{V}_\mathrm{D}$, and class-aware dense prompt embedding $\mathbf{V}_\mathrm{S,D}$. (b) Visualization of class-specific fusion prompt embedding $\mathbf{V}_\mathrm{S,D}^i$, where $i$ corresponds to channels labeled 1–6 in the figure, representing both the channel and category indices. Pseudo-colored images $I\text{1}$, MVD $I_\text{2}$, and corresponding labels are shown above.}
\label{fig:output}
\end{figure*}
\subsection{Ablation Studies on Module Design}

We perform comprehensive module ablation studies on both datasets to assess the efficacy of the modules. As shown in Table \ref{AS1}, we first define the baseline model as the original SAM that takes the pseudo-colored SAR image as the input and only train the mask decoder part. The following models, M1 to M5, all take two inputs. Compared to our method PolSAM (w/ MVD), M2, M3, and M1 are designed to validate the effectiveness of the adapter, FFP, and their combined impact, respectively. It can be seen that the adapter in the PolSAM encoder has a significant influence of the model performance, and the proposed FFP can also improve the result to a certain extent. To validate the adapters' role in bridging the modality gap, Fig. \ref{fig::adapter} visualizes the encoder's output features and their t-SNE distributions. Compared to the baseline, which exhibits limited category discriminability and partial cluster overlap, the adapter-equipped encoder generates highly distinct feature activations and semantically separable clusters, thereby confirming successful modality alignment. The effectiveness of the proposed progressive SFP can be verified by comparing the performance of M4 and M5. Compared to M4, our method enhances performance by 3.97\% on the PhySAR-Seg-1 dataset, while on the PhySAR-Seg-2 dataset, the enhancement is 5.93\%. The specific contribution of SFP-2 is isolated by M5: removing it alone reduces the mIoU by 3.45\% and 4.94\% on the two datasets, respectively. Mechanistically, SFP-1 generates the initial prompts solely from the input embeddings, while SFP-2 refines them into class-aware ones by interacting with the high-level semantic features $\mathbf{F}_{\mathrm{att}}$ from the mask decoder. Consistently, the dedicated ablation visualization in Fig.~4 of the supplementary material shows that removing SFP-2 degrades segmentation details in ambiguous regions.


To further demonstrate the validity of our proposed method, we conducted an ablation study on the MVD. In the PolSAM (w/o MVD) experiment, we replaced the MVD with pseudo-colored images. Despite this modification, PolSAM (w/o MVD) still shows significant improvements over the baseline, achieving mIoU increases of 15.35\% on the PhySAR-Seg-1 dataset and 8.64\% on the PhySAR-Seg-2 dataset, illustrating the soundness of our overall model design. However, when compared to PolSAM (w/ MVD), PolSAM (w/o MVD) falls short. PolSAM (w/ MVD) achieves even greater mIoU increases of 17.06\% on PhySAR-Seg-1 and 9.89\% on PhySAR-Seg-2 compared to the baseline, highlighting that the MVD, representing the scattering mechanism, provides valuable multidimensional information that enhances segmentation performance, particularly in capturing details.

To further validate our frozen encoder design, we compared it against full (100\%) and partial (75\%, 50\%, 25\%) fine-tuning strategies without adapters (Table \ref{tab:finetune}). Results indicate that fine-tuning significantly increases trainable parameters (up to 94.1M) while simultaneously degrading segmentation accuracy. This degradation is primarily due to severe overfitting on the relatively small PolSAR datasets. By freezing the encoder and utilizing lightweight adapters (14.0M parameters), PolSAM successfully achieves the optimal balance between parameter efficiency and robust performance.

\begin{table}
\caption{Ablation Study on Encoder Fine-tuning Strategies}
\label{tab:finetune}
\centering
\resizebox{\linewidth}{!}{
\begin{tabular}{lccccccc}
\toprule
\rowcolor{gray!38}
\cellcolor{white} & \cellcolor{white} & \multicolumn{3}{c}{\textbf{PhySAR-Seg-1 (\%)}} & \multicolumn{3}{c}{\textbf{PhySAR-Seg-2 (\%)}}\\
\cmidrule(r){3-5} \cmidrule(l){6-8}
\rowcolor{gray!20}
\multirow{-2}{*}{\cellcolor{white}\textbf{Strategy}} & \multirow{-2}{*}{\cellcolor{white}\textbf{Params}} & \textbf{mAcc} & \textbf{mF1} & \textbf{mIoU} & \textbf{mAcc} & \textbf{mF1} & \textbf{mIoU} \\
\midrule
T (100\%) & 94.1M & 79.51 & 80.07 & 67.51 & \underline{75.55} & 57.21 & 43.40 \\
T (75\%)  & 71.5M & 82.60 & 83.35 & 71.91 & 72.49 & 56.34 & 42.47 \\
T (50\%)  & 50.2M & \underline{82.78} & \underline{83.60} & \underline{72.42} & 74.73 & \underline{59.18} & \underline{45.10} \\
T (25\%)  & 28.9M & 82.58 & 83.32 & 71.80 & 75.20 & 57.56 & 43.59 \\
\midrule
\textbf{Ours (F+A)} & \textbf{14.0M} & \textbf{83.70} & \textbf{84.71} & \textbf{73.55} & \textbf{76.47} & \textbf{61.40} & \textbf{46.55} \\
\bottomrule
\multicolumn{8}{l}{\footnotesize *T: Trainable layers in the image encoder; F+A: Frozen image encoder + Adapter.} \\
\end{tabular}
}
\end{table}

\subsection{Discussions on Semantic prompt}

\noindent \textbf{The Effectiveness of Semantic Prompts.}
To verify the effectiveness of the proposed SFP, we analyze the learned ultimate sparse and dense prompt embeddings at the semantic level on the PhySAR-Seg-1 dataset, namely $\mathbf{P}_\mathrm{S}^{(2)} \in \mathbb{R}^{N \times 256}$ and $\mathbf{P}_\mathrm{D}^{(2)} \in \mathbb{R}^{32 \times 32 \times 256}$, where $N$ denotes the number of sparse prompts. These embeddings reflect semantic prompts associated with class information and mask information, respectively. To better illustrate the semantic prompt, the dense prompt embedding $\mathbf{P}_\mathrm{D}^{(2)}$ is averaged along the channel dimension, yielding $\mathbf{V}_\mathrm{D} \in \mathbb{R}^{32 \times 32}$ for visualization. As shown in the second row of Fig. \ref{fig:output} (a), $\mathbf{V}_\mathrm{D}$ clearly highlights the semantic map corresponding to the annotated mask. Additionally, we multiply the sparse prompt embedding $\mathbf{P}_\mathrm{S}^{(2)}$ by the dense prompt embedding $\mathbf{P}_\mathrm{D}^{(2)}$ and average along the channel dimension to obtain the class-aware dense prompt visualization map $\mathbf{V}_\mathrm{S,D}$. Notably, the class-aware prompt reveals more discriminative semantics, as illustrated in the third row of Fig. \ref{fig:output} (a).

In our experiments, the number of sparse prompt $C$ is set to the number of semantic class. Consequently, we visualize $C$ class-specific fusion prompt maps denoted as $\mathbf{V}_\mathrm{S,D}^i, i=1,...,C$ in Fig. \ref{fig:output} (b) for discussion. It can be observed that the class-specific fusion prompt map shows significant correlation with each class. The results illustrate the explicit semantic meanings of the learned fusion prompt, which play an important role to guide the decoder to learn the segmentation result.

\begin{table}
    \centering
    \caption{Ablation study on the number of sparse prompts (SP)}
    \label{tab:5}
        \renewcommand{\arraystretch}{0.80}
    \begin{tabular}{lccc}
         \toprule
  \rowcolor{gray!38} No. of SP &mAcc &mF1 score &mIoU \\
         \cmidrule(r){1-4}
        3 &83.40 &84.07 &72.68 \\
        \textbf{6} (classes)  &\textbf{83.70} &\textbf{84.71} &\textbf{73.55} \\
        2 * 6 &82.34 &83.12 &71.42 \\
        4 * 6 &\underline{83.43} &\underline{84.29} &73.01  \\
        10 * 6 &83.07 &83.78 &72.25 \\
        20 * 6 &83.37 &84.28 &\underline{73.06}  \\
        \bottomrule
    \end{tabular}
\end{table}
\noindent \textbf{Sparse Prompt Parameter Selection.} 
The number of sparse prompts $N$ reflects the degree of semantic correlation. To evaluate its impact on model performance, we conducted ablation experiments using six different values of $N$, each being a multiple of the dataset's number of categories. As shown in Table \ref{tab:5}, the best performance on the PhySAR-Seg-1 dataset was achieved when $N=6$, matching the number of classes. Increasing $N$ to 10 or 20 times the number of categories increased the model's parameter count without yielding consistent performance improvements. Theoretically, matching $N$ to the number of classes enforces a one-to-one semantic alignment, where each prompt serves as a distinct information bottleneck. Over-parameterizing with a larger $N$ breaks this alignment, causing the model to capture spurious speckle noise and leading to parameter redundancy, while a value of $N$ smaller than the number of classes can cause semantic confusion. These results suggest that sparse prompts, when aligned with semantic information, facilitate effective semantic-level fusion, improving segmentation accuracy.

\begin{table}[htbp]
    \centering
    \renewcommand{\arraystretch}{0.70}
        \centering
        \caption{Ablation experiment on physical information input forms in the PhySAR-Seg-2 dataset}
        \label{Data2_A}
        \begin{tabular}{c|cccccc}
          \toprule
          \rowcolor{gray!38} Metrics& \multicolumn{6}{c}{Inputs: H/A/Alpha($\alpha$)\cite{cloude1997entropy}  T\cite{chen2018polsar}} \\
          \cmidrule(lr){2-7}
          \rowcolor{gray!20} (\%)  &None & H/A/$\alpha$ & T(6) & T(9) & H/A/$\alpha$\_T & MVD \\
          \midrule
          BK &29.53 & \underline{33.93} & 33.74 & 32.20 & 32.68 & \textbf{34.83} \\
          Montain & 64.25 & \underline{74.96} & 66.87 & 63.28 & 65.59 & \textbf{75.88} \\
          Vegetation & 13.23 & \underline{19.90} & 17.81 & 10.34 & 15.02 & \textbf{22.63} \\
          Developed & 9.21 &\textbf{23.12} & 15.48 & 18.21 & \underline{19.73} & 18.13 \\
          I-urban & 0.01 & 15.12 & 15.57 & \underline{19.66} & 11.93 & \textbf{30.76} \\
          H-urban & 18.98& \underline{31.68} & 26.21 & 20.56 & 31.30 & \textbf{42.43} \\
          L-urban & 30.65& \underline{46.83} & 43.43 & 42.59 & 42.36 & \textbf{54.57} \\
          Water & 90.51 & 91.81 & \textbf{94.15} & 93.55 & 93.74 & \underline{93.31} \\
          \hdashline
          mAcc  & 63.48& \underline{72.81} & 69.29 & 65.11 & 69.51 & \textbf{76.47} \\
          mF1 score  & 43.29 & \underline{57.34} & 56.02 & 51.59 & 52.82 & \textbf{61.40} \\
          mIoU  &32.05 &  \underline{42.25} & 39.14 & 37.53 & 39.05 & \textbf{46.55} \\
          \hdashline
          FPS & 26.63& \textbf{22.42} & 12.81 & 7.20 & 5.68 & \underline{18.74} \\
          Memory(GB) &- & \underline{6.49} & 12.99 & 19.46 & 25.9 & \textbf{0.049} \\
          \bottomrule
        \end{tabular}
\end{table}

\subsection{Discussions on Effectiveness of MVD.}
Utilizing MVD offers two notable benefits. Initially, the complex information is transformed into a compact format through a series of processing stages. This process allows the data to visually represent the scattering mechanism, effectively conveying the physical information in a way that is intuitively understandable for humans, while also being highly correlated with semantic information. Additionally, MVD requires only a small amount of memory, significantly reducing the storage needed for complex data and improving efficiency in data loading and preprocessing for neural networks.

To validate the effectiveness of MVD, we conducted experiments comparing different polarimetric decomposition features as substitutes for MVD in the PolSAM. The results, shown in Table \ref{Data2_A}, evaluate the performance of four feature combinations, primarily based on the H/A/Alpha decomposition method \cite{cloude1997entropy} and the polarization coherence matrix (\( T \)).  For detailed information, refer to the appendix. H/A/Alpha is a three-channel feature, and \( T \) represents the complete polarization information in PolSAR imaging \cite{chen2018polsar}. The real parts of the upper triangular elements of \( T \) form the six-channel T(6) dataset, as shown in the third column of Table \ref{Data2_A}. The full set of real and imaginary components together constitute the nine-channel T(9) dataset, represented in the fourth column. Additionally, the H/A/Alpha\_T(12) approach in the fifth column combines the three channels from the H/A/Alpha decomposition with the nine-channel data from the \( T \) decomposition, resulting in a comprehensive twelve-channel dataset. We compared the performance metrics, frame per second (FPS), and data memory usage against these alternative decomposition methods.

The comparative results in Table primarily validate the efficacy of the PolSAM architecture, as all four variants incorporating polarimetric information outperform the baseline. More importantly, the superiority of our MVD representation is evident. When benchmarked against the second-best method (H/A/Alpha decomposition), our approach yields substantial gains, improving mAcc by 3.66\%, mean F1 score by 4.06\%, and mIoU by 4.3\%. These margins clearly demonstrate that integrating MVD provides a richer feature representation, significantly enhancing the segmentation capability of the PolSAM framework.

To provide deeper insights, we conducted a class-wise IoU analysis on the PhySAR-Seg-2 dataset (Table \ref{Data2_A}). MVD achieves the best IoU in six of the eight categories. The advantage is most pronounced in the three urban classes characterized by double-bounce scattering and low sample proportions, where MVD surpasses the best alternative feature by 11.10\% (I-urban), 10.75\% (H-urban), and 7.74\% (L-urban), while remaining comparable on the stable water class (93.31\% vs. 94.15\%). This highlights the model's robustness in capturing complex scattering mechanisms even in data-scarce scenarios.

For loading a batch of 668 image pairs, our method reduces the loading time by 16.52 seconds, 57.11 seconds, and 82.07 seconds compared to the T(6), T(9), and H/A/Alpha\_T(12) approaches, respectively. In terms of data storage, MVD occupies only 49 MB of memory, representing just 0.755\%, 0.377\%, 0.252\%, and 0.189\% of the capacity utilized by the other four methods. These results highlight the clear advantages of MVD, making it a highly efficient and effective option for polarimetric data processing.

\section{Conclusion}
In this study, we proposed PolSAM, a novel segmentation framework designed for PolSAR image analysis. PolSAM utilizes the Microwave Vision Data (MVD) representation, which encodes scattering characteristics into a lightweight and interpretable format, offering both visual and physical insights. By integrating features from pseudo-colored SAR images and MVD, PolSAM effectively combines complementary information to guide the segmentation process. The framework incorporates a FFP module for the initial integration of features, supported by an adapter to align representation spaces and address feature differences. Additionally, the SFP module refines segmentation through multi-level feature interactions, enabling the effective incorporation of semantic and physical scattering characteristics. Together, these components enable PolSAM to achieve efficient and accurate segmentation while maintaining high interpretability.

In the future, we aim to explore more robust methods to enhance the adaptability of large models to SAR data for improved utilization efficiency. Furthermore, considering that the physical scattering characteristics inherent in PolSAR data are closely related to the spatial structures of ground targets, a promising future direction is extending our physical-visual fusion paradigm from 2D segmentation to advanced 3D perception tasks \cite{gao2026cosurfgs, fei2025knowledge}.
\small
\bibliographystyle{IEEEtran}
\bibliography{IEEEabrv,paper_cite}

\begin{IEEEbiography}[{\includegraphics[width=1in,height=1.25in,clip,keepaspectratio]{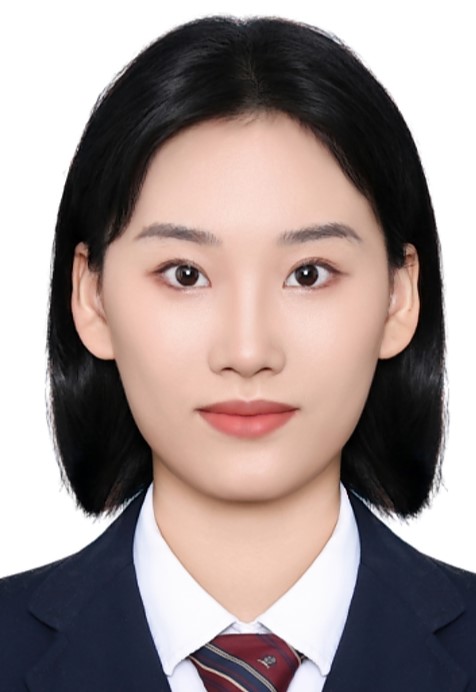}}]{Yuqing Wang} is currently working toward the Ph.D. degree in the School of Automation, Northwestern Polytechnical University, Xi'an, China. Her research interests include computer vision, synthetic aperture radar (SAR) image interpretation, multimodal image fusion, medical image processing, and adaptation of foundation models. \end{IEEEbiography}

\begin{IEEEbiography}[{\includegraphics[width=1in,height=1.25in,clip,keepaspectratio]{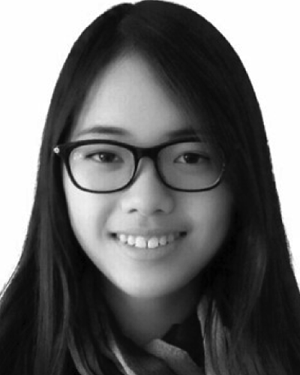}}]{Zhongling Huang} (huangzhongling@nwpu.edu.cn) received her B.Sc. degree in electronic information science and technology from Beijing Normal University, Beijing, China, in 2015, and her Ph.D. degree from the University of Chinese Academy of Sciences (UCAS) and the Aerospace Information Research Institute, Chinese Academy of Sciences, Beijing, China, in 2020. She served as a visiting scholar in the German Aerospace Center during 2018–2019, funded by UCAS. She is currently working in the Brain and Artificial Intelligence Lab, School of Automation, Northwestern Polytechnical University, Xi’an 710072, China. Her research interests include explainable deep learning for synthetic aperture radar (SAR), SAR image interpretation, deep learning, and remote sensing data mining.\end{IEEEbiography}

\vspace{-1.5cm}

\begin{IEEEbiography}[{\includegraphics[width=1in,height=1.25in,clip,keepaspectratio]{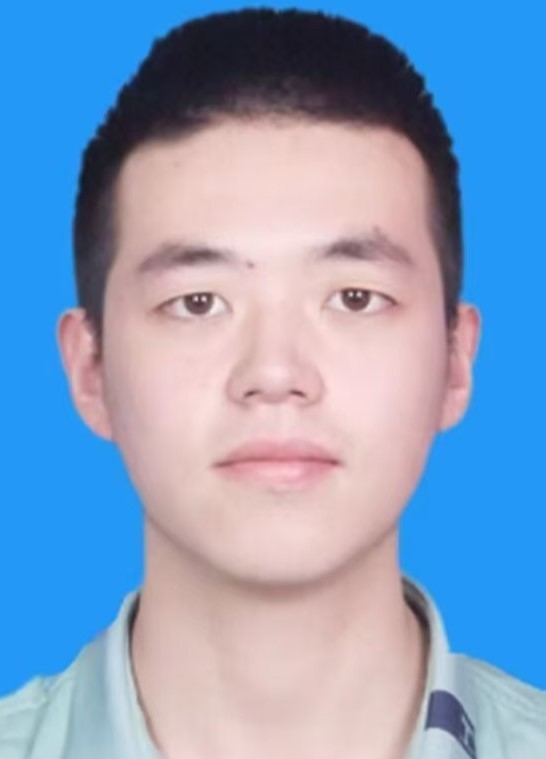}}]{Shuxin Yang} is currently a master's student at Northwestern Polytechnical University in Xi'an, Shanxi. His research direction is SAR image target detection in computer vision.\end{IEEEbiography}

\vspace{-1.5cm}

\begin{IEEEbiography}
[{\includegraphics[width=1in,height=1.25in,clip,keepaspectratio]{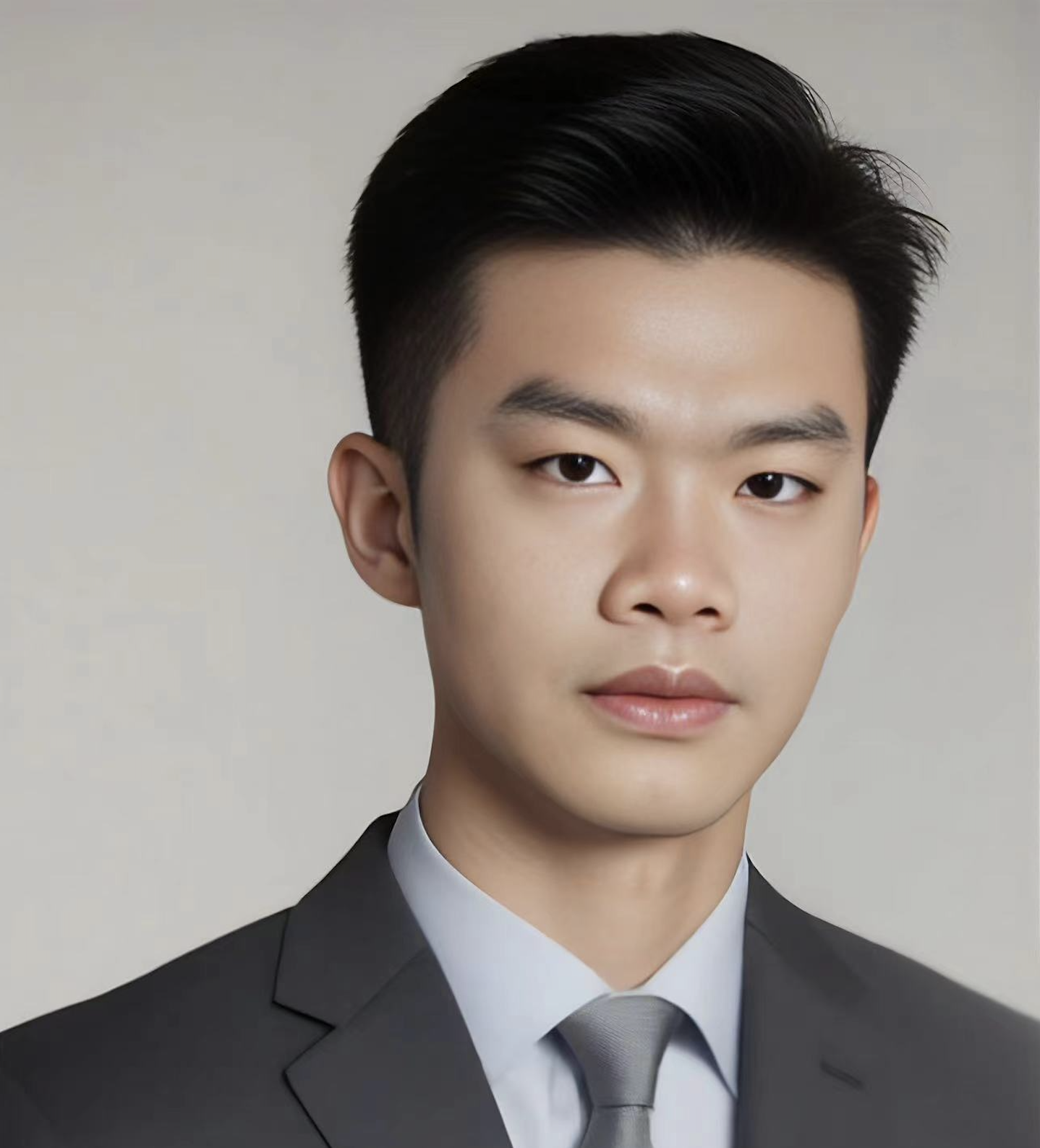}}]{Hao Tang} is an Assistant Professor at Peking University, China. Previously, he held postdoctoral positions at both CMU, USA, and ETH Zürich, Switzerland. He earned a master's degree from Peking University, China, and a Ph.D. from the University of Trento, Italy. He was a visiting Ph.D. student at the University of Oxford, UK, and an intern at IIAI, UAE. His research interests include machine learning, computer vision, embodied AI, and AIGC (including LLM), as well as their applications in scientific domains. \end{IEEEbiography}

\vspace{-1.5cm}

\begin{IEEEbiography}[{\includegraphics[width=1in,height=1.25in,clip,keepaspectratio]{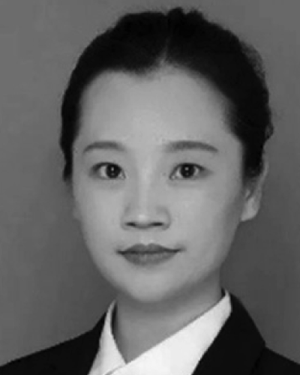}}]{Xiaolan Qiu} (Senior Member, IEEE) received the B.S. degree in electronic engineering and information science from the University of Science and Technology of China, Hefei, China, in 2004, and the Ph.D. degree in signal and information processing from the Graduate University of Chinese Academy of Sciences, Beijing, China, in 2009. She is with the National Key Laboratory of Microwave Imaging Technology, Chinese Academy of Sciences, Beijing 100090, China, and also with the Aerospace Information Research Institute, Chinese Academy of Sciences, Beijing 100094, China. Her research interests include synthetic aperture radar (SAR) imaging, SAR simulation, and SAR 3-D reconstruction.\end{IEEEbiography}

\vspace{-1.5cm}

\begin{IEEEbiography}[{\includegraphics[width=1in,height=1.25in,clip,keepaspectratio]{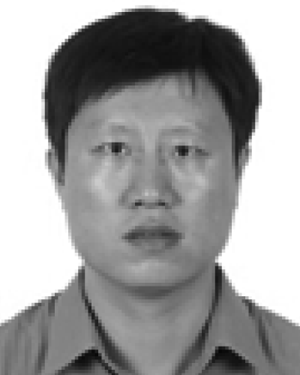}}]{Junwei Han} (Fellow, IEEE) is currently a Professor in the School of Automation, Northwestern Polytechnical University. His research interests include computer vision, pattern recognition, remote sensing image analysis, and brain imaging analysis. He has published more than 70 papers in top journals such as IEEE TPAMI, TNNLS, IJCV, and more than 30 papers in top conferences such as CVPR, ICCV, MICCAI, and IJCAI. He is an Associate Editor for several journals such as IEEE TNNLS and IEEE TMM.\end{IEEEbiography}

\vspace{-1.5cm}

\begin{IEEEbiography}[{\includegraphics[width=1in,height=1.25in,clip,keepaspectratio]{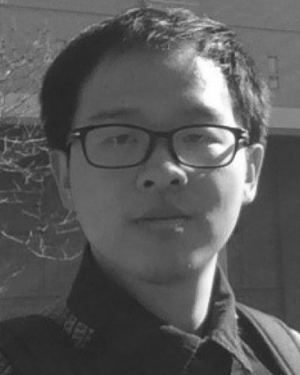}}]{Dingwen Zhang} (Member, IEEE) received the Ph.D. degree from Northwestern Polytechnical University, Xi’an, China, in 2018. He is a Professor with the School of Automation, Northwestern Polytechnical University. From 2015 to 2017, he was a Visiting Scholar with the Robotic Institute, Carnegie Mellon University, Pittsburgh, PA, USA. His research interests include computer vision and multimedia processing, especially on saliency detection and weakly supervised learning.\end{IEEEbiography}

\end{document}